\pdfoutput=1
\documentclass[letterpaper, 10 pt, conference]{ieeeconf}  
\usepackage{times}
\usepackage{graphicx}
\usepackage{amsmath}
\usepackage{amssymb}
\usepackage{subcaption}
\usepackage{multirow}
\usepackage{hyperref}
\usepackage{color}

\IEEEoverridecommandlockouts                              
\title{\LARGE \bf
InterFace: Adjustable Angular Margin Inter-class Loss for Deep Face Recognition}

\begin{document}

\author{Meng Sang$^{1,2}$, Jiaxuan Chen$^{2,3}$, Mengzhen Li$^{1,2}$
,Pan Tan$^{1,2}$, Anning Pan$^{1,2}$, Shan Zhao$^{1,2}$, Yang Yang$^{1,2}$\\
$^{1}$Yunnan Normal University, 650500, Kunming, China\\
$^{2}$Laboratory of Pattern Recognition and Artificial Intelligence, 650500, Kunming, China\\
$^{3}$Zhejiang University, 310058, Hangzhou, China \\
 Emails: {sangmeng.one@gmail.com, yyang\_ynu@163.com}
}
\pagestyle{plain}
\maketitle

\begin{abstract}
	In the field of face recognition, it is always a hot research topic to improve the loss solution to make the face features extracted by the network have greater discriminative power. 
	Research works in recent years has improved the discriminative power of the face model by normalizing softmax to the cosine space step by step and then adding a fixed penalty margin to reduce the intra-class distance to increase the inter-class distance.
	Although a great deal of previous work has been done to optimize the boundary penalty to improve the discriminative power of the model, adding a fixed margin penalty to the depth feature and the corresponding weight is not consistent with the pattern of data in the real scenario.
	To address this issue, in this paper, we propose a novel loss function, InterFace, releasing the constraint of adding a margin penalty only between the depth feature and the corresponding weight to push the separability of classes by adding corresponding margin penalties between the depth features and all weights.
	To illustrate the advantages of InterFace over a fixed penalty margin, we explained geometrically and comparisons on a set of mainstream benchmarks. 
	From a wider perspective, our InterFace has advanced the state-of-the-art face recognition
	performance on five out of thirteen mainstream benchmarks. All training codes, pre-trained models, and training logs, are publicly released.
	\footnote{$https://github.com/iamsangmeng/InterFace$}.
\end{abstract}

\section{Introduction}
\label{sec:intro}
With the development of face recognition technology, it has been applied to various fields in life, such as finance, security, and enterprises. 
The face recognition system consists of the process of image acquisition, face detection, face alignment, feature extraction, and feature matching.
In the process of face matching, the vectors generated in the feature extraction are required to measure the similarity with all the faces. 
Knowing that it is intuitive that the model should have small intra-class distances for samples of the same identity and large inter-class distances for  samples of different identities. Hence, in the face system, we are required to make the right decision boundary, even if the picture of the face changes dramatically under the same identity, and also to reject the imposter under a different identity.\par

Although the accuracy of face recognition has improved greatly, it has not yet achieved the expected results \cite{kim2020groupface}. 
Most of the recent related studies  \cite{schroff2015facenet,sohn2016improved,chopra2005learning,liu2016large,liu2017sphereface,wang2018cosface,deng2019arcface,huang2020curricularface,jiao2021dyn,duan2019uniformface,meng2021magface,boutros2022elasticface} have focused on improving the loss function. 
The content of our work focuses on the problem of fixed margin penalty existing in the direction of Classification Task \ref{ct} and we propose our our solution based on ArcFace \cite{deng2019arcface}.
Next, we will sort out the related work and problems in previous loss studies and summarize them in two directions for metric learning and classification tasks as follows.\par

\paragraph{Deep Learning}
In the direction of metric learning, the design of losses  \cite{schroff2015facenet,sohn2016improved,chopra2005learning} is based on triplet. 
Facenet  \cite{schroff2015facenet}  directly learns a mapping from face images to a compact Euclidean space where distances directly correspond to a measure of face similarity.
N-pair  \cite{sohn2016improved} proposed objective function firstly generalizes triplet loss by allowing joint comparison among more than one negative examples – more specifically, $N-1$ negative examples – and secondly reduces the computational burden of evaluating deep embedding vectors via an efficient batch construction strategy using only $N$ pairs of examples, instead of $(N+1)×N$. 
Contrastive  \cite{chopra2005learning} losses learn a function that maps input patterns into a target space such that the $L_1$ norm in the target space approximates the "semantic" distance in the input space. 
However, the number of triplets explodes during the training period as the number of samples in the training dataset increases.
\par
\paragraph{Classification Task}
\label{ct}
In the direction of the classification task, subsequent losses \cite{liu2016large,liu2017sphereface,wang2018cosface,deng2019arcface,huang2020curricularface,jiao2021dyn,duan2019uniformface,meng2021magface,boutros2022elasticface} are designed on the basis of softmax losses.
Liu et al. \cite{liu2016large} proposed a large-margin softmax (L-Softmax) by introducing penalty margin ideas for softmax to encourage intra-class compactness and inter-class separability between learned features.
SphereFace \cite{liu2017sphereface} extends previous work on L-Softmax by further constraining the weights of fully connected layers to impose discriminative constraints on a hypersphere manifold, which intrinsically matches the prior that faces also lie on a manifold. SphereFace deploys a multiplicative angular penalty margin between the deep features and their corresponding weights.
In CosFace \cite{wang2018cosface}, it is proposed to add a cosine angle between depth features and weights. CosFace fixes the norm value of the depth feature and the corresponding weight and proposes to scale the norm of the depth feature to a constant s.
ArcFace \cite{deng2019arcface} proposed additive angular margin by deploying angular penalty margin on the angle between the deep features and their corresponding weights. 
The great success of softmax loss with penalty margin motivated several works to propose a novel variant of softmax loss.
All these solutions achieved notable accuracies on mainstream benchmarks for face recognition. 
Huang et al. \cite{huang2020curricularface} propose an Adaptive Curriculum Learning loss (CurricularFace) that embeds the idea of curriculum learning into the loss function to achieve a novel training strategy for deep face recognition, which mainly addresses easy samples in the early training stage and hard ones in the later stage. 
In Dyn-ArcFace \cite{jiao2021dyn}, the traditional fixed additive angular margin is developed into a dynamic one, which can reduce the degree of overfitting caused by the fixed additive angular margin. 
Duan et al. \cite{duan2019uniformface} propose Learning Deep Equidistributed Representation for Face Recognition that imposes an equidistributed constraint by uniformly spreading the class centers on the manifold.
Meng et al. \cite{meng2021magface} propose A Universal Representation for Face Recognition and Quality Assessment called MagFace that learns a general embedding feature whose dimension measures the quality of the given image. In this way, the embedded features of the face can be regularly distributed around the class center according to their dimensions.
In Boutros1's work, ElasticFace \cite{boutros2022elasticface} refuted the assumption of uniform distribution of class centers made by ArcFace and CosFace, and proposed an elastic margin loss for deep face recognition called ElasticFace.
However, adding a fixed margin penalty to the depth feature and the corresponding weight creates a fixed decision boundary between the depth feature to the other weights. This is not consistent with the data patterns in real scenarios.  
\par

To address the issues mentioned in the Classification Task direction \ref{ct}, in this work, we propose InterFace loss. InterFace first introduces the idea of adding different penalty margins to different class centers for a single sample to solve the problem of adding a fixed margin penalty to different class centers for a single sample. On this basis, there are the sample-to-class center distance and inter-class distance of corresponding class centers to other class centers ratios(SIR) are introduced to generate a penalty margin for different class centers by a specified function.
To demonstrate the geometric space advantage of our method, we provide a geometric interpretation of the decision boundary and provide a toy example of a simple network we customize to implement a simple 8-classification problem.
At the same time to illustrate the effectiveness of our InterFace loss on face recognition accuracy, we will report on 13 benchmarks. We will compare and analyze the effect achieved by the whole method with the recent state-of-the-art. In a more detailed comparison, with prior work and recent state-of-the-art, our InterFace continuously extends state-of-the-art face recognition performance on five benchmarks. Especially the age-related datasets \cite{moschoglou2017agedb,zheng2017cross} such as AgeDB-30 and CALFW have better performance.\par

We summarize the contributions as follows:
\begin{itemize}
	\item We redefine ArcFace loss by decision boundary equivalence and propose a novel InterFace loss to improve the discriminative power of the network.
	
	\item We experimentally demonstrate that the redefined ArcFace and the original ArcFace losses are equivalent. We do extensive ablation and toy example of geometric interpretation to illustrate the superiority of InterFace.
	
	
	\item We have done a lot of ablation experiments in the field of face recognition, and in the final face verification benchmark, 5 benchmarks have been extended and the others are very close to the top state-of-the-art performance.
\end{itemize}
\par
\section{InterFace Loss}
We  propose in this work a novel learning loss stragegy, InterFace loss, aiming at imporving the accuracy of face recognition by optimizing spatial distribution between and within classes. Different from previous work that  constrain a same margin value from the deep feature to the corresponding weight, our proposed InterFace loss  introduce SIR to constrain the margin values from the deep feature to the other weights. The  margin values are generated by SIR through a convex function.
\paragraph{Softmax Loss}
Softmax is a normalized exponential function that is used for the output of the last fully-connected layer and is often 
used with the cross entropy function as the most widely used classification loss function. Softmax is defined as follows:
\begin{equation}
	\begin{aligned}
		L_{softmax}(x_i,y_i)
		&= -\frac{1}{N} \sum_{i=1}^{N} \log {\frac{e^{f_{y_i}}}{\sum_{j=1}^{c}e^{f_j}}}
		\\
		&=-\frac{1}{N} \sum_{i=1}^{N} \log \frac{e^{x_iW_{y_i}^T} + b_{y_i}}{\sum_{j=1}^{c}e^{x_iW_J^T + b_j}}
	\end{aligned}
\end{equation}
where $x_i \in R^d$ is the depth feature extracted by Deep Convolutional Neural Networks(DCNNs) of sample $z_i$ and  belongs to $y_i$ class ($z_i$ integer in the range $[1,n]$ and $y_i$ integer in the range $[1,c]$). Given that $n$ is the number of samples and $c$ is number of classes. The embedding feature dimension d is set 512 by this paper following  \cite{deng2019arcface,wang2018cosface,boutros2022elasticface}. $W_{y_i}$ represents $y_i$ column of  weights $W \in R_c^d $ and $b_{y_i}$ represents the corresponding bias offset. The batch size is represented by $n$.\par
In a simple binary class classification, assuming that input $z_i$ belongs to class 1, the model will correctly classify $z_i$ if $W_1^Tx_i+b1 >  W_2^Tx_i+b2$ and $z_i$ will be classified as class 2 if $W_2^Tx_i+b2 >  W_1^Tx_i+b1$. Therefore the decision boundary of softmax loss is $x(W_1^T - W_2^T) +b1 - b2 =0 $. However, the softmax loss function cannot explicitly optimize the embedded features to minimize intra-class distance and maximize inter-class distance. To address the limitations of softmax, the idea of decision boundary introduces softmax and then a series of optimization works based on angular margin penalty. This idea has always guided our improvement in loss and is also the most popular loss function for face recognition model training.\par
\paragraph{Angular Margin Penalty-based Loss}
Following  \cite{liu2016large,liu2017sphereface,wang2018cosface}, the bias offset, for simplicity, can be fixed to $b_{y_i}$ = 0. In the above case, the function $f_{yi}$ can be simplified as $x_iW_{y_i}^T= \left\|x_i\right\| \left\|W_{y_i}\right\|\cos(\theta_{y_i})$, where $\theta_{y_i}$ is the angle between the $y_i$ column of the weights of the last fully-connected layer $W$ and the embedding feature $x_i$. By fixing the weights norm and the feature norm to $\left\| W_{y_i}\right\|=1$ and $\left\|x_i\right\| = 1$, respectively, and rescaling the $x_i$ to constant s, the output of the softmax activation function is subject to the cosine of the angle $\theta_{y_i}$. Following  \cite{liu2017sphereface,wang2018cosface,deng2019arcface}, this type of loss has a margin penalty under fixed conditions, we call this type of loss function a fixed angular margin penalty loss function(AML). 
The AML is defined as follows:
\begin{small}
	\begin{equation}	
		L_{AML}  = 
		-\frac{1}{N} \sum_{i=1}^{N} \log \frac{e^{s(\cos(m_1\theta_{y_i}+m_2)-m_3)}}
		{e^{s(\cos(m_1\theta_{y_i}+m_2)-m_3)}\!+\! \sum\limits_{j=1,j \ne y_i}^{c}e^{s(\cos(\theta_j))}}
		\label{equation2}
	\end{equation}
\end{small}
\begin{figure*}
	\centering
	\begin{subfigure}{0.49\linewidth}
		\centering
		\label{fig:db_arc}
		\includegraphics[width=1\textwidth]{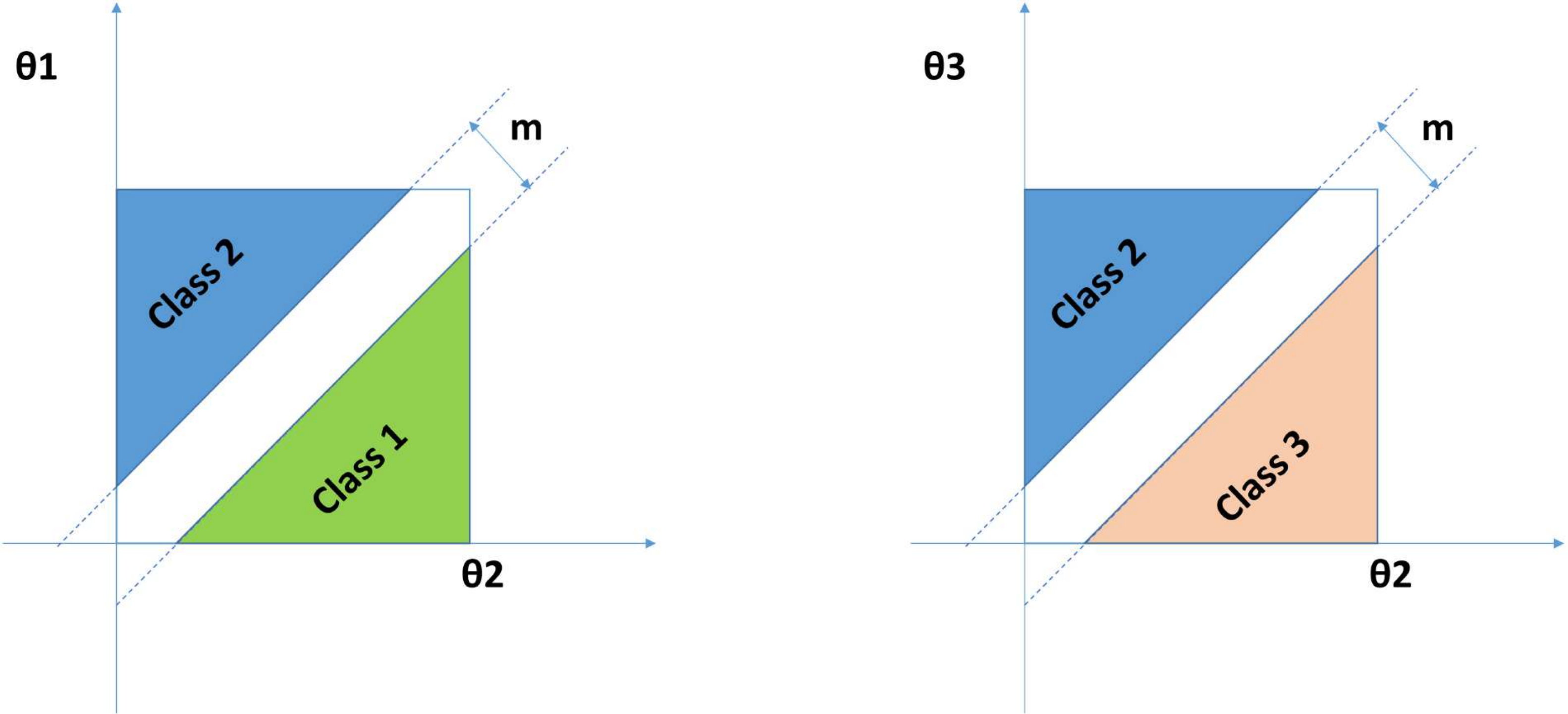}
		\caption{ArcFace}
	\end{subfigure}
	\hfill
	\begin{subfigure}{0.49\linewidth}
		\centering
		\label{fig:db_iti}
		\includegraphics[width=1\textwidth]{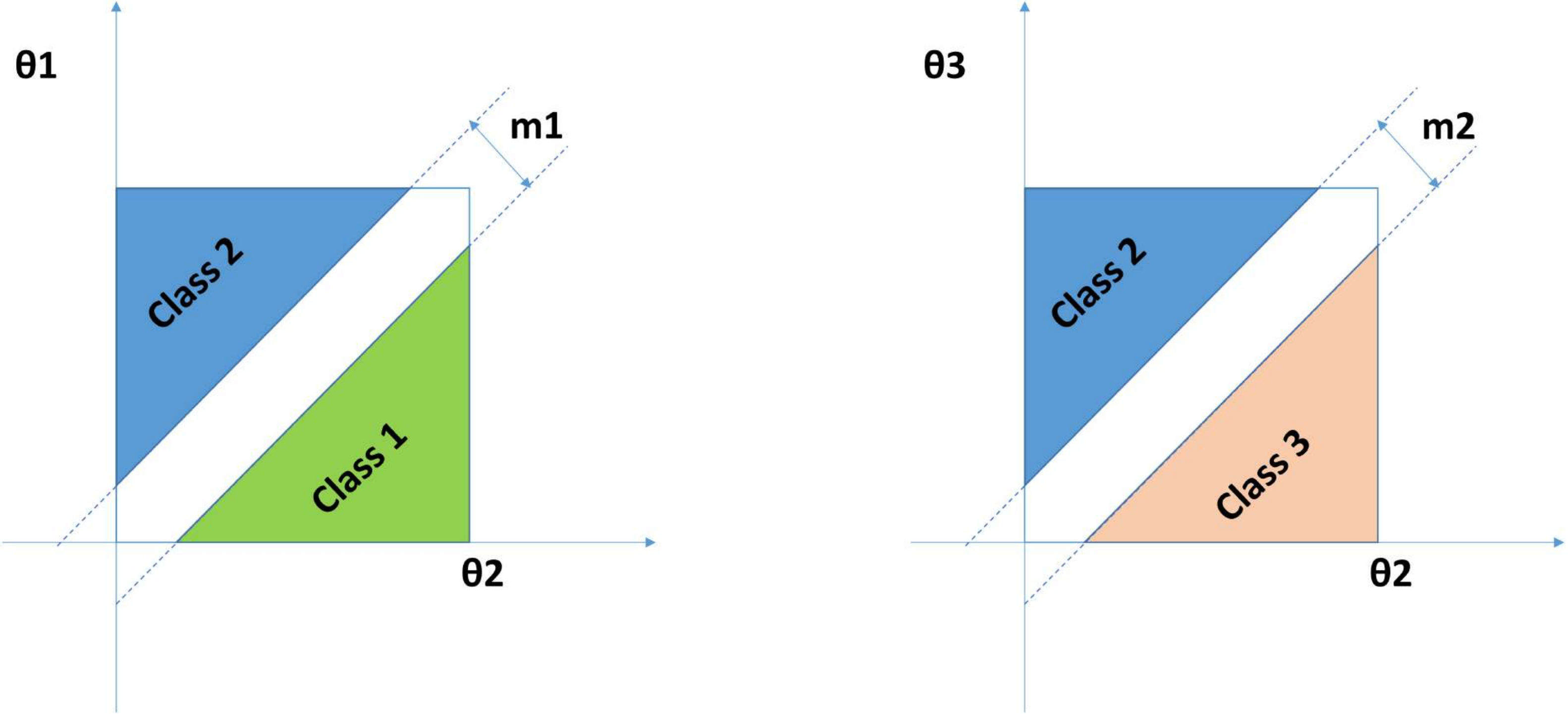}
		\caption{InterFace}
	\end{subfigure}
	\caption{Dcesion boundary of (a) ArcFace and (b) InterFace for multi-classification. The dashed blue line is the decision boundary. The white gray illustrates the decision margin. In (a) arcface, class 2 adds decision boundary of the same size as class 1 and class 2. In (b) InterFace, class 2 adds the same small decision boundary to class 1 and class 3. In InterFace, class 2 adds decision boundary of different sizes to class 1 and class 3.
	}
	\label{fig:db}
\end{figure*}
where $m_1$, $m_2$ and $m_3$ are the margin penalty parameters proposed by SphereFace \cite{liu2017sphereface}, ArcFace \cite{deng2019arcface}, and CosFace \cite{wang2018cosface}, respectively. Under different values of $m_1$, $m_2$, and $m_3$, different decision boundaries will be obtained and different feature representations needed for face verification will be optimized. \par
When $m_1=$1, $m_2=1$, $m_3=0$, this is the operation loss without adding any margin angle optimization. In the previous example, assuming that the input $z_i$ belongs to label 1, when $\cos(\theta_1) > \cos(\theta_2)$ condition is true, the label can be correctly classified.
Therefore, the decision margin of this loss is $\cos(\theta_1) - \cos(\theta_2) = 0$. However, this faces the same problem as the aforementioned softmax in Euclidean space, thus motivating research work on adding margin angles. In SphereFace, by setting $m_1=\alpha ( \alpha > 1.0)$, $m_2=0$ and $m_3=0$, the problem of imposing margin angle penalty. The decision  boundary  of SphereFace is than $\cos(\theta_{y_i}) - \cos(\theta_j)  = 0 $.
Cosface \cite{wang2018cosface} proposed cosine margin penalty by setting $m_1=1$, $m_2=0$, and $m_3 = \alpha ( 0 < \alpha < 1 - \cos(\frac{\pi}{4}) )$. The decision boundary of CosFace is  $ \cos(\theta_{y_i}) - \cos(\theta_j) -m_3 = 0$. Arcface \cite{deng2019arcface} proposed additive angular margin penalty by setting $m_1=1$, $m_2= \alpha ( 0<\alpha<1.0 )$ and $m_3 = 0$. The decision boundary of Arcface is than $\cos(\theta_{y_i} + m_2) - \cos(\theta_j)= 0$\par
In ArcFace \cite{deng2019arcface}, CosFace \cite{wang2018cosface} and SphereFace \cite{liu2017sphereface}, they all use the important concept of margin
penalty to optimize and experiment the optimal penalty margin values. 
But in all the previous works, the margin angle added by one sample to other classes at the same time is fixed. This does not correspond to the actual distribution of classes in space. Therefore, this motivated us to propose InterFace to sample the distribution of the model space.
\par
\paragraph{InterFace}
The proposed InterFace loss is extended over the angular margin penalty-based loss by adding different margin penalties for other sample centers. First, define an equation as follows:
\begin{equation}	
	\begin{aligned}
		m = m_1 + m_2 
		\label{formul6}
	\end{aligned}
\end{equation}
where $m,m_1$, and $m_2$ are three real numbers satisfying the equality rule. We let the imposed margin penalty be $m$, and according to ArcFace in Equation \ref{equation2}, we can know that its decision boundary  is $\cos(\theta_{y_i} + m) - \cos(\theta_j) = 0$. According to equation (6), we can transform the Arcface decision boundary into $\cos(\theta_{y_i} + m_1) - \cos(\theta_j - m_2) = 0$.  We redefine ArcFace losses(RArc) as follows:
\begin{small}
	\begin{equation}	
		\begin{aligned}
			L_{RArc} 
			& = -\frac{1}{N} \sum_{i=1}^{N} \log \frac{e^{s(\cos(\theta_{y_i}+m_1)}}
			{e^{s\cos(\theta_{y_i}+m_1)} + \sum_{j=1,j \ne y_i}^{c}e^{s(\cos(\theta_j - m_2))}}
			\label{formul7}
		\end{aligned}
	\end{equation}
\end{small}
\begin{table}
	\centering
	\scalebox{0.65}{
		\begin{tabular}{|c|c|c|c|c|c|c|c|c|c|c|c|c|}
			\hline
			\multirow{2}*{$m_1$}&\multirow{2}*{$m_2$} & \multicolumn{2}{|c|}{LFW} & \multicolumn{2}{|c|}{AgeDB-30} &\multicolumn{2}{|c|}{CALFW} &\multicolumn{2}{|c|}{CPLFW} &\multicolumn{2}{|c|}{CFP-FP} & \multirow{2}*{BC sum}	\\
			\cline{3-12}
			~		 & ~ 		& Acc(\%)& BC 	&Acc(\%)		& BC    & Acc(\%)	& BC & Acc(\%)	& BC   	&Acc(\%) 	& BC 	& ~ \\
			\hline
			0.5		 & 0 		& 99.22 & 1		& 93.7 			&4 		& 93.13 	& 2  & 88.63 	& 3 	& 94.79 	& 3 	& 13 \\
			\hline
			0.4 	 & 0.1		& 99.32 & 4 	& 93.62 		& 3 	& 93.02 	& 1  & 88.52 	& 1 	& 94.8 		& 4 	& 13 \\
			\hline
			0.3 	 & 0.2 		& 99.23 & 4 	& 93.6 			& 2 	& 93.23 	& 4  & 88.6 	& 2 	& 94.79 	& 3 	& 13 \\
			\hline
			0.2 	 & 0.3 		& 99.32 & 4 	& 93.35 		& 1 	& 93.18 	& 3  & 88.67 	& 4 	& 94.64 	& 1 	& 13 \\
			\hline 
		\end{tabular}
	}
	\caption{
		Validation for equation \ref{formul7}. When $m=0.5$, the enumeration has specific values that conform to equation \ref{formul6}, and experiments are performed according to equation \ref{formul7}. The Board Count (BC) was reported for the whole experiment. Correctness of equation \ref{formul7} evaluates using BC sum. In all training, the used architecture is ResNet50 trained on CASIA \cite{yi2014learning}.
	}
	\label{tab:mm}
\end{table}
By using the experimental results in Table.\ref{tab:mm}, we can show that ArcFace and redefine ArcFace losses are equivalent under the evaluation of the Bount Count sum(BC sum). The previous experiments provide support for our proposal of InterFace. To add different margin penalties for other sample centers, We introduce  concepts of SIR and define a function $\gamma$ as follows:
\begin{equation}	
	\begin{aligned}
		\gamma(z_i,W_{y_i},W_j) 
		& = \frac{d_{s}}{d_{inter}}
		\\
		& = \frac{\arccos(\left\|z_i\right\| \left\|W_{y_i}^T\right\|)}{\arccos(\left\|W_{y_i}\right\| \left\|W_{j}^T\right\|)}
	\end{aligned}
\end{equation}
and threshold function $t$ define as follows:
\begin{equation}	
	\begin{aligned}
		t(d_{inter})= 
		&a + \frac{b}{d_{inter}}
		\\
		&= a +  \frac{b}{\arccos(\left\|W_{y_i}\right\| \left\|W_{j}^T\right\|)}
	\end{aligned}
\end{equation}
where $d_{s}$ is angle between the sample vector $z_i$ that belong $y_i$ and the sample center $W_{y_i}$ and  ${d_{inter}}$ is angle between sample center $W_{y_i}$ and $W_{j}(j \ne y_i)$.  $t$ is a function of $d_{inter}$. We define a function on the exponent to control the margin to the different class centers, the function is as follows:
\begin{equation}	
	\begin{aligned}
		f(\gamma,t)
		& = \alpha(\exp^{-\gamma} - \exp^{-t})
	\end{aligned}
\end{equation}
where $\alpha$ a is to control the size of the boundary float. Further, we can get about InterFace Loss (IF) as follows:
\begin{small}
	\begin{equation}	
		\begin{aligned}
			L_{IF} 
			& = -\frac{1}{N} \sum_{i=1}^{N} \log \frac{e^{s(\cos(\theta_{y_i}+m)}}
			{e^{s(\cos(\theta_{y_i}+m)} + \sum_{j=1,j \ne y_i}^{c}e^{s(\cos(\theta_j - f(\gamma,t)))}}
		\end{aligned}
	\end{equation}
\end{small}
where we use $f(\gamma,t)$ to control the margin penalty between the current sample and different classes. The decision boundary of InterFace are $\cos(\theta_{y_i} + m)  - \cos(\theta_j - f(\gamma,t)) = 0$. Figure.\ref{fig:db} illustrates the decision boundary of ArcFace and InterFace for different inter-classes. This aims at optimizing the sample distribution of the feature space. \par

In order to select the appropriate parameters for the experiments, we designed the above method into three sets of experiments as follows.
In the first experimental design we set the fixed inter-class distance to $d_{inter} = \frac{pi}{2}$ and set the hyperparameter $b=0$ according to the uniform distribution in  \cite{deng2019arcface}, and obtain the values of $\alpha$ and $a$ through the experiment, which we named InterFace$_{cid\&ct}$.
In the second experimental design we use the real class spacing $d_inter$ and set the hyperparameter $b=0$ and experiment to obtain the values of $a$ and $b$, We name this experiment InterFace$_{did\&ct}$
The last set of experimental design we use the real class spacing $d_{inter}$ and set hyperparameter $a=0$ and get the value of b by experiment, we named this experiment InterFace$_{did\&dt}$.
We have designed three groups of experiments and a large number of ablation experiments to demonstrate the validity of the parameters and to report the final comparative results.
\par

\paragraph{Parameter Selection}
\label{ps}
Parameter Selection is to select the optimal parameters of different experiments.
In order to choose the optimal parameter value, we choose CASIA \cite{yi2014learning} as our training dataset and Resent50 \cite{he2016deep} as our training model.
For eavlation, we use the sum of Board count on LWF \cite{huang2008labeled}, AgeDB-30 \cite{moschoglou2017agedb}, CALFW \cite{zheng2017cross},CPLFW \cite{zheng2018cross} and CFP-FP \cite{sengupta2016frontal} as our parameter selected evaluation criteria.
The current model of the selected parameters and the subsequent training models are improved based on arcface-torch in Insightface\footnote{$https://github.com/deepinsight/insightface/tree/master/\\recognition/arcface\_torch$}.

In the experiment of InterFace$_{cid\&ct}$, we use the control variable method to obtain the values of the parameters $\alpha$ and $a$. First of all, we set the value of $d_{inter}$ to $\frac{\pi}{2}$, the value of $b$ to 0 , and the value of $\alpha$ to 0.1 where $d_{inter}$ is based on the angles of adjacent vectors under the premise of uniform distribution\cite{deng2019arcface}, and we select an optimal $a$ value that we enumerate the values of $[0.1,0.6]$ and the step size is 0.1 for experiments. By observing our experimental results, the best experimental result $a$ is observed to be 0.2 (in Tabel.\ref{tab:ccm}). Further, we can choose the optimal $\alpha$ parameter by setting the value of $a$ to 0.2 for the control variable. The range of the enumerated values $[0.06,0.14]$ and the step size is 0.02. It can be seen by observing the experimental results that 0.1 is the best experimental result observed (in Tabel.\ref{tab:cca}).
Therefore, based on the experimental results above, we set the final parameters of experiment InterFace$_{cid\&ct}$ to be $d_{inter}=\frac{\pi}{2}$, $a=0.2$, $b=0$, and $\alpha=0.1$.
\begin{table}
	\centering
	\scalebox{0.65}{
		\begin{tabular}{|c|c|c|c|c|c|c|c|c|c|c|c|c|}
			\hline
			\multirow{2}*{$\alpha$}&\multirow{2}*{$a$} & \multicolumn{2}{|c|}{LFW} & \multicolumn{2}{|c|}{AgeDB-30} &\multicolumn{2}{|c|}{CALFW} &\multicolumn{2}{|c|}{CPLFW} &\multicolumn{2}{|c|}{CFP-FP} & \multirow{2}*{BC sum}	\\
			\cline{3-12}
			~		 & ~ 		& Acc(\%)& BC 	&Acc(\%)		& BC    & Acc(\%)	& BC & Acc(\%)	& BC   	&Acc(\%) 	& BC 	& ~ \\
			
			\hline
			0.1 & 0.1 & 99.433 & 5 & 94.867 & 5 & 93.850 & 6 & 89.300 & 2 & 95.057 & 3 &  21 \\
			\hline
			0.1 & 0.2 & 99.450 & 6 & 95.033 & 6 & 93.483 & 2 & 89.567 & 6 & 95.214 & 5 & \textbf{25} \\
			\hline
			0.1 & 0.3 & 99.383 & 3 & 94.517 & 2 & 93.600 & 3 & 89.350 & 3 & 95.243 & 6 & 17 \\
			\hline
			0.1 & 0.4 & 99.367 & 2 & 94.950 & 4 & 93.600 & 4 & 89.500 & 5 & 95.071 & 4 & 19 \\
			\hline 
			0.1 & 0.5 & 99.333 & 1 & 94.650 & 3 & 93.517 & 1 & 89.417 & 4 & 94.943 & 2 & 11  \\
			\hline 
			0.1 & 0.6 & 99.400 & 4 & 94.317 & 1 & 93.600 & 5 & 89.183 & 1 & 94.929 & 1 & 12 \\
			\hline 
		\end{tabular}
	}
	\caption{
		Parameter selection for $a$ of InterFace$_{cid\&ct}$. The Board count (BC) is used to evaluate the quality of the model under the specified parameter and The final $a$ is selected based on the BC sum. 
	}
	\label{tab:ccm}
\end{table}

\begin{table}
	\centering
	\scalebox{0.65}{
		\begin{tabular}{|c|c|c|c|c|c|c|c|c|c|c|c|c|}
			\hline
			\multirow{2}*{$\alpha$}&\multirow{2}*{$a$} & \multicolumn{2}{|c|}{LFW} & \multicolumn{2}{|c|}{AgeDB-30} &\multicolumn{2}{|c|}{CALFW} &\multicolumn{2}{|c|}{CPLFW} &\multicolumn{2}{|c|}{CFP-FP} & \multirow{2}*{BC sum}	\\
			\cline{3-12}
			~		 & ~ 		& Acc(\%)& BC 	&Acc(\%)		& BC    & Acc(\%)	& BC & Acc(\%)	& BC   	&Acc(\%) 	& BC 	& ~ \\
			\hline
			0.06 & 0.2 & 99.400 & 2 & 94.767 & 3 & 93.800 & 4 & 89.167 & 2 & 95.371 & 5 & 14  \\
			\hline
			0.08 & 0.2 & 99.317 & 1 & 94.800 & 4 & 93.633 & 2 & 89.233 & 3 & 95.086 & 1 & 11 \\
			\hline
			0.10 & 0.2 & 99.450 & 3 & 95.033 & 5 & 93.483 & 1 & 89.567 & 5 & 95.214 & 3 & 17  \\
			\hline
			0.12 & 0.2 & 99.467 & 5 & 94.533 & 1 & 93.650 & 2 & 89.550 & 4 & 95.286 & 4 & 16 \\
			\hline 
			0.14 & 0.2 & 99.450 & 3 & 94.717 & 2 & 93.850 & 5 & 89.017 & 1 & 95.100 & 2 &  12 \\
			\hline 
		\end{tabular}
	}
	\caption{Parameter selection for $\alpha$ of InterFace$_{cid\&ct}$. The Board count (BC) is used to evaluate the quality of the model under the specified parameter and The final $\alpha$ is selected based on the BC sum.}
	\label{tab:cca}
\end{table}
In the experiment of InterFace$_{did\&ct}$, the result that $\alpha$ and $a$ are related is shown by Tabel.\ref{tab:ccm} and Tabel.\ref{tab:cca}. Therefore, we set $\alpha$ to 0.1 and  enum $a$ of values that the range is $[0.1,0.6]$ and step is 0.1 to find the optimal parameters. By observing our experimental result Tabel.\ref{tab:dcm}, we select parameters of the highest BC sum to complete the next training. 
Therefore, based on the experimental results above, we set the final parameters of experiment InterFace$_{did\&ct}$ to be $a=0.3$, $b=0$, and $\alpha=0.1$.\par
\begin{table}
	\centering
	\scalebox{0.65}{
		\begin{tabular}{|c|c|c|c|c|c|c|c|c|c|c|c|c|}
			\hline
			\multirow{2}*{$\alpha$}&\multirow{2}*{$a$} & \multicolumn{2}{|c|}{LFW} & \multicolumn{2}{|c|}{AgeDB-30} &\multicolumn{2}{|c|}{CALFW} &\multicolumn{2}{|c|}{CPLFW} &\multicolumn{2}{|c|}{CFP-FP} & \multirow{2}*{BC sum}	\\
			\cline{3-12}
			~		 & ~ 		& Acc(\%)& BC 	&Acc(\%)		& BC    & Acc(\%)	& BC & Acc(\%)	& BC   	&Acc(\%) 	& BC 	& ~ \\
			\hline
			0.1 & 0.1 						& 99.433 & 4 & 94.550 & 2 & 93.767 & 3 & 89.267 & 4 & 95.200 & 5 & 18  \\
			\hline
			0.1 & 0.2 						& 99.367 & 2 & 94.600 & 4 & 93.600 & 1 & 89.033 & 1 & 94.943 & 2 & 10 \\
			\hline
			0.1 & 0.3 						& 99.450 & 6 & 94.850 & 5 & 93.817 & 5 & 89.383 & 6 & 94.857 & 1 & 23 \\
			\hline
			0.1 & 0.4 						& 99.433 & 4 & 94.383 & 1 & 93.633 & 2 & 89.200 & 3 & 95.143 & 4 & 14\\
			\hline 
			0.1 & 0.5 						& 99.350 & 1 & 94.583 & 3 & 93.867 & 6 & 89.067 & 2 & 95.114 & 3 & 15 \\
			\hline 
			0.1 & 0.6 						& 93.367 & 2 & 94.950 & 6 & 93.767 & 3 & 89.350 & 5 & 95.286 & 6 & 22 \\
			\hline 
		\end{tabular}
	}
	\caption{
		Parameter selection for $a$ of InterFace$_{did\&ct}$. The Board count (BC) is used to eavaluate the quailty of the model under the specified parameter and The final $a$ are selected based on BC sum.
	}
	\label{tab:dcm}
\end{table}
In the experiment of InterFace$_{did\&dt}$, the result that $\alpha$ and $a$ are related is shown by Tabel.\ref{tab:ccm} and Tabel.\ref{tab:cca}. Therefore, we set $\alpha$ to 0.1 and enum $b$ of values that the range is $[10,60]$ and step is 10 to find the optimal parameters. By observing our experimental result Tabel.\ref{tab:ddm}, we select parameters of the highest BC sum to complete the next training. 
Therefore, based on the experimental results above, we set the final parameters of experiment InterFace$_{did\&dt}$ to be $a=0$, $b=10$, and $\alpha=0.1$.\par
\begin{table}
	\centering
	\scalebox{0.65}{
		\begin{tabular}{|c|c|c|c|c|c|c|c|c|c|c|c|c|}
			\hline
			\multirow{2}*{$\alpha$}&\multirow{2}*{$b$} & \multicolumn{2}{|c|}{LFW} & \multicolumn{2}{|c|}{AgeDB-30} &\multicolumn{2}{|c|}{CALFW} &\multicolumn{2}{|c|}{CPLFW} &\multicolumn{2}{|c|}{CFP-FP} & \multirow{2}*{BC sum}	\\
			\cline{3-12}
			~		 & ~ 		& Acc(\%)& BC 	&Acc(\%)		& BC    & Acc(\%)	& BC & Acc(\%)	& BC   	&Acc(\%) 	& BC 	& ~ \\
			\hline
			0.1 & 10 						& 99.450 & 6 & 94.733 & 5 & 93.900 & 6 & 89.200 & 2 & 95.086 & 4 & 23  \\
			\hline
			0.1 & 20						& 99.367 & 2 & 94.617 & 2 & 93.650 & 1 & 89.117 & 1 & 94.857 & 1 & 7 \\
			\hline
			0.1 & 30 						& 99.433 & 4 & 94.633 & 3 & 93.667 & 2 & 89.367 & 3 & 95.200 & 5 & 17 \\
			\hline
			0.1 & 40 						& 99.400 & 3 & 94.283 & 1 & 93.683 & 3 & 89.600 & 5 & 95.257 & 6 & 18 \\
			\hline 
			0.1 & 50 						& 99.433 & 4 & 94.933 & 6 & 93.850 & 5 & 89.450 & 4 & 94.871 & 2 & 21 \\
			\hline 
			0.1 & 60 						& 99.317 & 1 & 94.700 & 4 & 93.733 & 4 & 89.617 & 6 & 95.000 & 3 & 18 \\
			\hline 
		\end{tabular}
	}
	\caption{
		Parameter selection for $b$ of InterFace$_{did\&dt}$. The Board count (BC) is used to evaluate the quality of the model under the specified parameter and The final $b$ is selected based on BC sum.
	}
	\label{tab:ddm}
\end{table}
\begin{figure*}
	\centering
	\begin{subfigure}{0.45\linewidth}
		\label{fig:te_arc}
		\includegraphics[width=1 \textwidth]{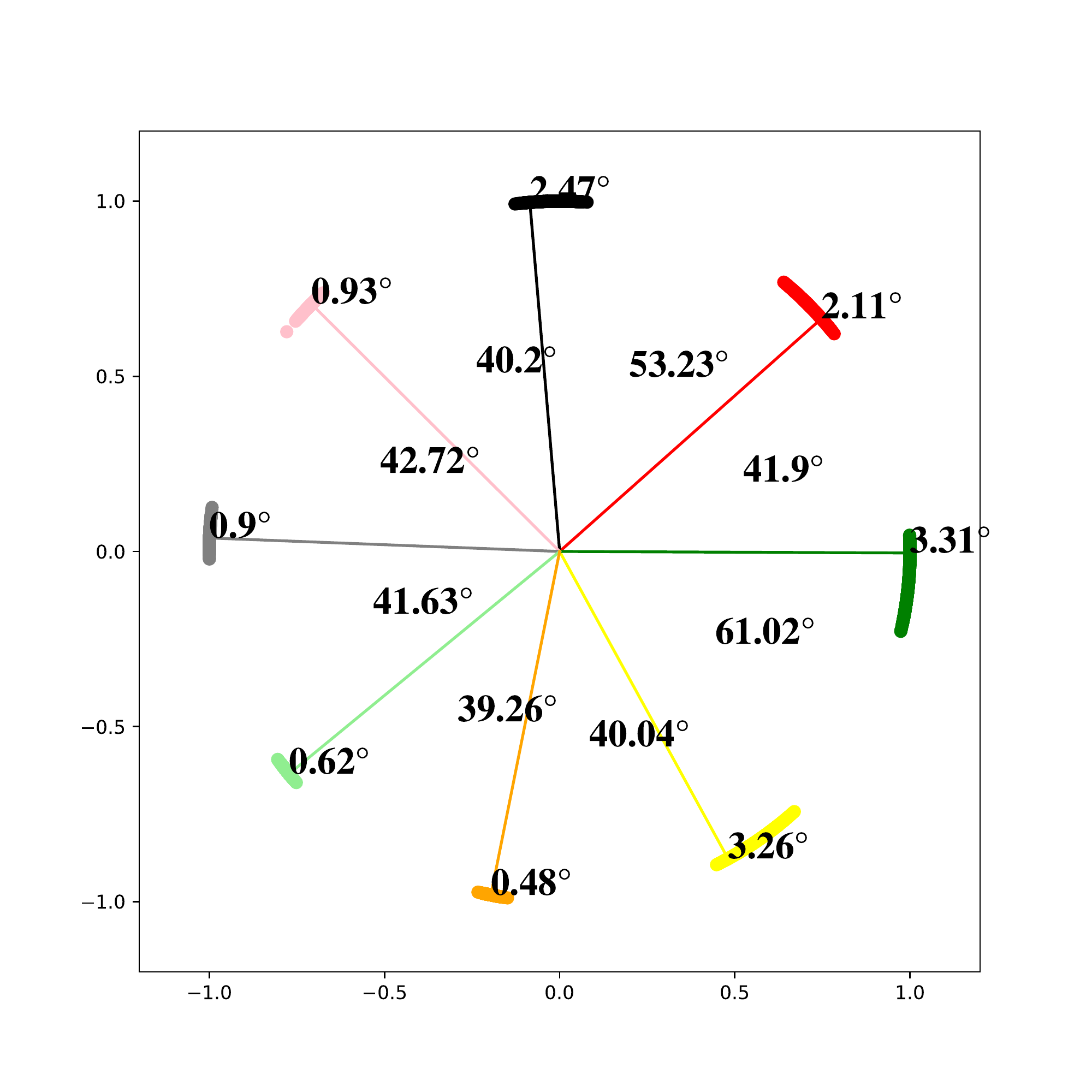}
		\caption{arcface}
	\end{subfigure}
	\begin{subfigure}{0.45\linewidth}
		\label{fig:te_cc}
		\includegraphics[width=1 \textwidth]{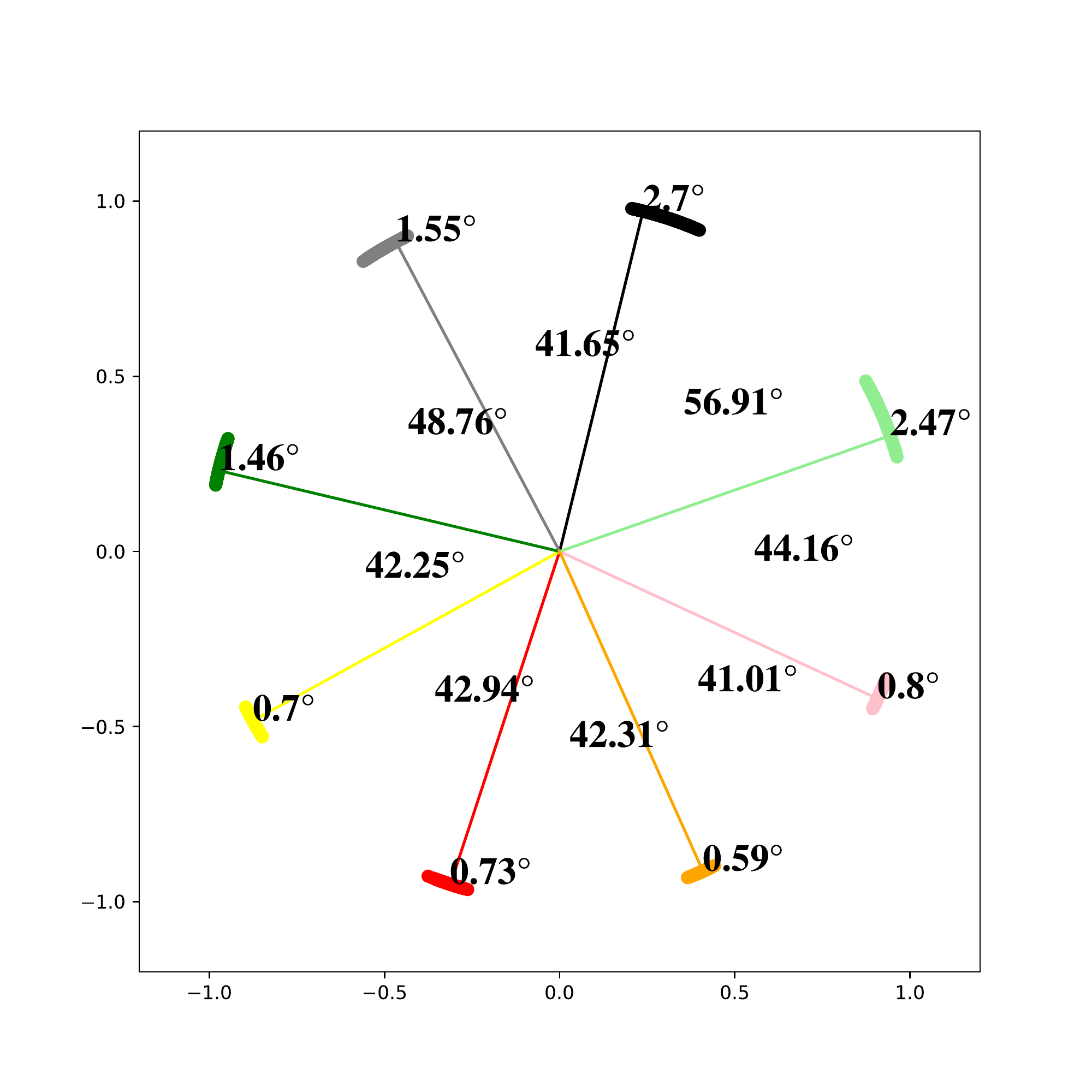}
		\caption{InterFace$_{cid\&ct}
			$}
	\end{subfigure}
	\begin{subfigure}{0.45\linewidth}
		\label{fig:te_dc}
		\includegraphics[width=1 \textwidth]{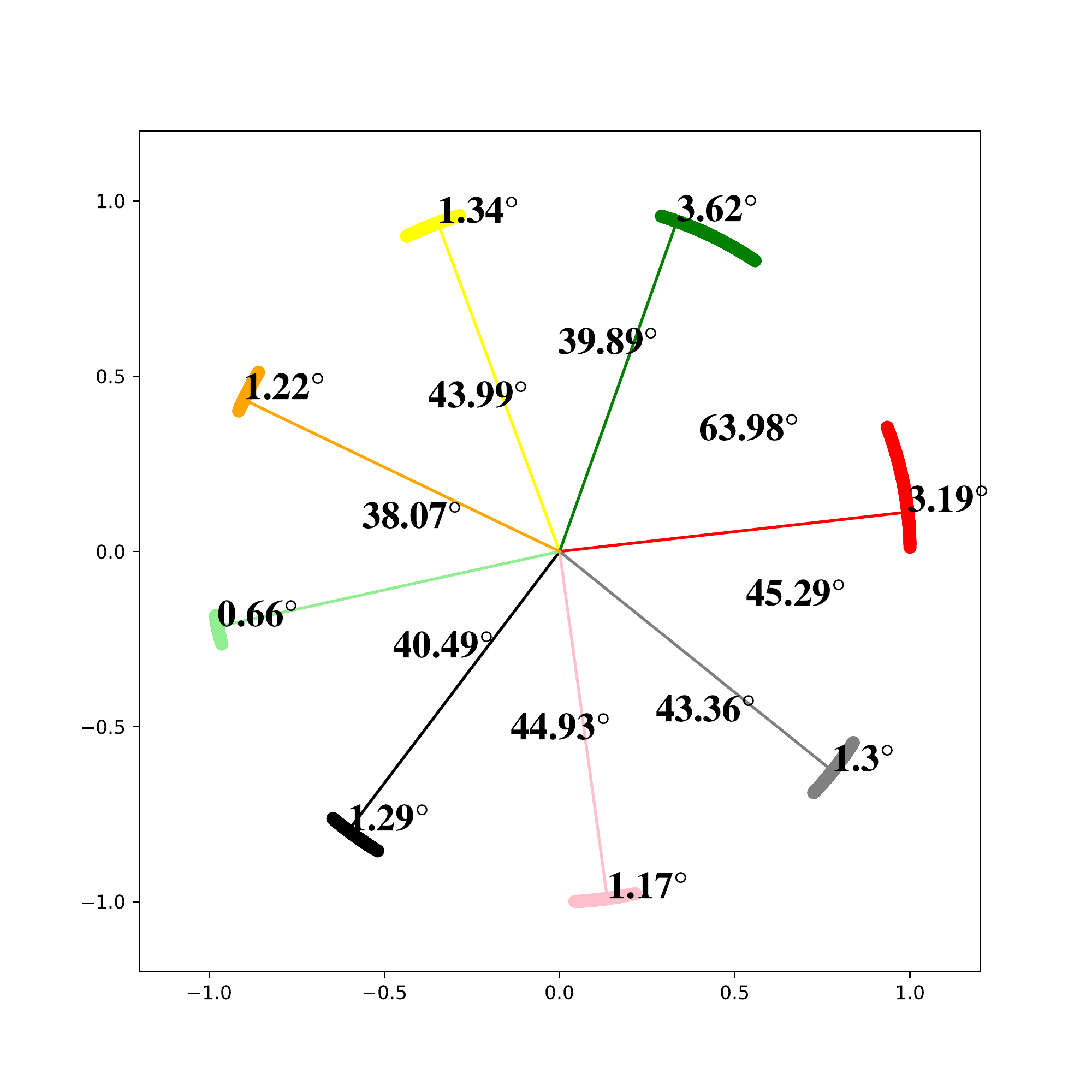}
		\caption{InterFace$_{did\&ct}
			$		}
	\end{subfigure}
	\begin{subfigure}{0.45\linewidth}
		\label{fig:te_dd}
		\includegraphics[width=1 \textwidth]{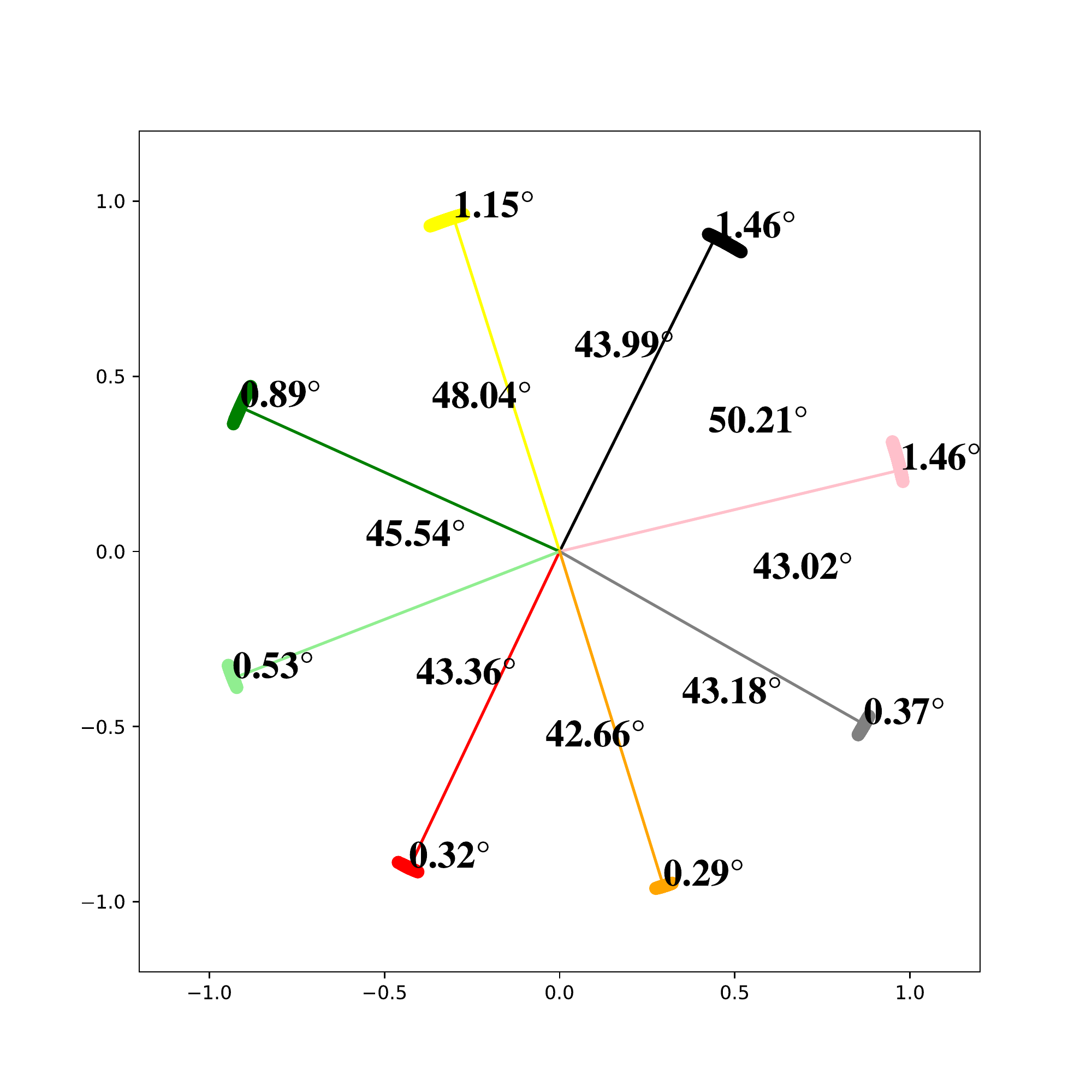}
		\caption{InterFace$_{did\&dt}
			$		}
	\end{subfigure}	
	\caption{
		Toy examples of arcface, InterFace$_{cid\&ct}$, InterFace$_{did\&ct}$, and InterFace$_{did\&dt}$ are trained under the optimal parameter. The 2-D Feature embeddings are normalized. Thus, the feature embeddings are allocated around the class centers in arcos space with a fixed radius. The inner number indicates the decision boundary between classes, and the outer number indicates the variance of the samples to the class center. A comparison of Arcface and InterFace shows that classes of InterFace are more evenly distributed in 2-D space.
	}
	\label{fig:te}
\end{figure*}
\paragraph{Toy example}
In order to illustrate that InterFace can better cluster and separate the feature space, we performed corresponding experiments on InterFace$_{cid\&ct}$, InterFace$_{dit\&ct}$ and InterFace$_{dit\&dt}$ respectively, and selected the corresponding parameters of the experiment according to the \ref{ps}, and showed the results of the experiment accordingly.
We built a toy example and used a custom network architecture to map samples of 8 different identities into 2-D feature embeddings. Since the MS1MV2-8\cite{boutros2022elasticface} training set has not been made public, we screened the corresponding MS1MV2-8 training set for training according to the protocol of the MS1MV2-8 training set. 
According to the experimental results in the Tabel.\ref{ps}, we set ($d_{inter}=\frac{pi}{4}$, $\alpha=0.1$, $a=0.2$, and $b=0$), ($\alpha=0.1$, $a=0.3$, and $b=0$) and ($\alpha=0.1$, $a=0$, and $b=10$) for InterFace$_{cid\&ct}$, InterFace$_{dit\&ct}$ and InterFace$_{dit\&dt}$ respectively. In the Figure.\ref{fig:te}, we show the parameters and experimental settings of arcface and the other three groups of experiments respectively. By comparing the experimental results, we can observe that InterFace samples between classes in the space are better distributed around the class centers and the class centers are more uniformly separated in the space.\par

\section{Experiments}
\paragraph{Training settings}
We use the Resnet100  \cite{he2016deep} network architecture to train our InterFace. This was motivated by the wide use of this architecture in the state-of-the-art face recognition solutions \cite{wang2018cosface,deng2019arcface,boutros2022elasticface}. We follow the common settings and set the scale parameter s to 64, Stochastic Gradient Descent(SGD) optimizer with an initial learning rate of 1e-1, set momentum parameter to 0.9, and weight decay parameter to 5e-4. The learning rate is divided by 10 at 80k, 140k, 210k, and 280k training iterations. The total number of training iterations is 295K. The model is implemented using the PyTorch framework. We set batch-size to 512 and train our model on Linux device (CentOS Linux release 7.9.2009) with Intel(R) Xeon(R) Silver 4216 CPU @ 2.10GHz, 128RAM and 2 Nvidia GeForce RTX 6000 GPUs. The network training and subsequent evaluation generate 512-d embedded features on images of size 3*112*112. During training, the images were randomly flipped horizontally with a probability of 0.5, and the data was normalized between -1 and 1.
\paragraph{Training dataset}
We follow the latest research work \cite{boutros2022elasticface} and choose to use the MS1MV2 dataset for training. The results of our comparison with state-of-the-art are shown in Section IV. MS1MV2 is a redefined version of MS-Celeb-1M  \cite{guo2016ms} in  \cite{deng2019arcface}, containing 5.8M images of 85k identities.
\paragraph{Evaluation benchmarks and metrics}
To illustrate the effectiveness of InterFace, we compare the experimental accuracy results with extensive methods in Section IV. We compared the 13 benchmarks below of 8 verification sets. First we report the validation accuracy using the unconstrained labeled field validation set protocols LFW \cite{huang2008labeled}, AgeDB-30 \cite{moschoglou2017agedb}, CALFW \cite{zheng2017cross}, CPLFW \cite{zheng2018cross} and CFP-FP \cite{sengupta2016frontal} protocols. With the rapid development of the face field in recent years, the accuracy of these validation sets has been saturated. Therefore we introduce the IJB-C \cite{maze2018iarpa} and IJB-B \cite{whitelam2017iarpa} validation sets and report when the true acceptance rates (TAR) at false acceptance rates (FAR) of  FAR is 1e-4, 1e-5, and 1e-6 respectively. MegaFace(Rank-1) \cite{kemelmacher2016megaface} and MageFace(R)(Rank-1) \cite{deng2019arcface} report our model performance with Rank-1 and true acceptance rates (TAR) at false acceptance rates (FAR) of 1e-6, respectively.
%
\begin{table*}
	\centering
	\scalebox{1.0}{
		\begin{tabular}{|c|c|c|c|c|c|c|}
			\hline
			\multirow{2}*{Method}& \multirow{2}*{Training Dataset} 			 & LFW 		 	 	&  Agedb\_30  	    & CALFW 	 		& CPLFW 	  		& CFP-FP \\
			\cline{3-7}
			~ 					 & 	~		   					 		 	 & Accuracy(\%)  	& Accuracy(\%)	    &Accuracy(\%)       &Accuracy(\%) 		&Accuracy(\%) \\
			\hline
			Arcface \cite{deng2019arcface}
			&    MS1MV2 \cite{deng2019arcface,guo2016ms}
			& 99.82(3) 	 	& 98.15     	    & 95.45     		& 92.08             &  98.27   \\
			CosFace \cite{wang2018cosface} 			 
			&    private      					 		 
			& 99.73         	& -			 	    & -					&  -		        & -	\\
			Dynamic-AdaCos \cite{zhang2019adacos}
			&  MS1MV2 \cite{guo2016ms,zhang2019adacos} + CASIA \cite{yi2014learning}			 			 
			&99.73 			&-				    & -					&  -		        & -	\\
			AdaptiveFace \cite{liu2019adaptiveface} 		 
			&    clean MS1M \cite{deng2019arcface,wu2018light} 							
			& 99.62         	& -			 	    & -					&  -		 	    & -				\\
			UniformFace \cite{duan2019uniformface}
			&  MS1MV2 \cite{deng2019arcface,guo2016ms}  + VGGFace2 \cite{cao2018vggface2} 			      	 
			& 99.8        	 	& -				    & -			 		&  -		        & -	\\
			GroupFace \cite{kim2020groupface}
			&  MS1MV2 \cite{deng2019arcface,guo2016ms} 								 
			& \textbf{99.85}(1)& 98.28		        & 96.20(3) 			& 93.17 		    & 98.63 \\
			CircleLoss  \cite{sun2020circle}			 
			& clean MS1M \cite{guo2016ms,sun2020circle}	
			& 99.73 			& -					& - 				& - 				& 96.02 \\
			CurricularFace \cite{huang2020curricularface} 		 
			& MS1MV2 \cite{deng2019arcface,guo2016ms}  									
			& 99.80			& 98.32				& 96.20(3) 			& 93.13				& 98.37 \\
			Dyn-arcFace \cite{jiao2021dyn} 		 
			& MS1MV2 \cite{deng2019arcface,guo2016ms}  									 
			& 99.80 			& 97.76 			& - 				& - 				& 94.25  \\
			MagFace \cite{meng2021magface} 			 
			& MS1MV2 \cite{deng2019arcface,guo2016ms} 
			& 99.83(2) 		& 98.17 			& 96.15 			& 92.87 			& 98.46 \\
			Partial-FC-ArcFace \cite{an2021partial}
			& MS1MV2 \cite{deng2019arcface,guo2016ms} 
			& 99.83(2) 		& 98.20 			& 96.18 			& 93.00 			& 98.45 \\
			Partial-FC-CosFace \cite{an2021partial}
			& MS1MV2 \cite{deng2019arcface,guo2016ms} 
			& 99.83(2) 		& 98.03 			& 92.20(3) 			& 93.10 			& 98.51 \\
			ElasticFace-Arc \cite{boutros2022elasticface}
			& MS1MV2 \cite{deng2019arcface,guo2016ms} 
			& 99.82(3) 		& 98.35(3) 			& 96.17 			& \textbf{93.28}(1) & 98.60(3) \\
			ElasticFace-Cos \cite{boutros2022elasticface}
			& MS1MV2 \cite{deng2019arcface,guo2016ms} 
			& 99.82(3) 		& 98.28 			& 96.18 			& 93.23 (2) 		& \textbf{98.73}(1) \\
			\hline
			InterFace$_{cid\&ct}$  & MS1MV2 \cite{deng2019arcface,guo2016ms} 
			& 99.83(2) 		& 98.37(2) 			& 96.22(2)			& 93.25(2) 			& 98.66(2) \\
			InterFace$_{did\&ct}$  & MS1MV2 \cite{deng2019arcface,guo2016ms} 
			& 99.80 			& 98.25 			& 96.12				& 93.10 			& 98.43 \\
			InterFace$_{did\&dt}$  & MS1MV2 \cite{deng2019arcface,guo2016ms} 
			& 99.83(2) 		& \textbf{98.38}(1) & \textbf{96.27}(1) & \textbf{93.28}(1) & 98.5 \\
			\hline
		\end{tabular}
	}
	\caption{The achieved results on the LFW, AgeDB-30, CALFW, CPLFW, and CFP-FP benchmarks.On large age gap(AgeDB-30) and Cross-Age LFW (CALFW), InterFace solutions extend state-of-the-art performances. InterFace scored very close to the state-of-the-art on LFW and CALFW. All decimal points are provided as reported in the respective works. The top performance in each benchmark is in bold. The top three performances in each benchmark are noted with a rank number between parentheses(1,2 or 3).}
	\label{tab:lfw}
\end{table*}

\section{Results}
Table.\ref{tab:lfw}, Table.\ref{tab:mega} and Table.\ref{tab:mm} present the results of this implementation on the 13 benchmarks. We mainly observe that our current proposed InterFace solution achieves state-of-the-art in 5 of the 13 benchmarks, and is also very close to the state-of-the-art in the other 8 benchmarks. For the fairness and rationality of the whole comparison environment, we select the work of the previous training environment under MS1MV2 \cite{deng2019arcface} or the redefined MS1MV2 training set, and our method InterFace is also trained under this dataset.
Comparing the accuracy results achieved by our proposed InterFace with the current state-of-the-art , we achieved the best results in AgeDB-30 \cite{moschoglou2017agedb}, CALFW \cite{zheng2017cross}, CPLFW \cite{zheng2018cross}, IJBB(Tar@Far=1e-5) \cite{whitelam2017iarpa} and IJBC(Tar@Far=1e-5) \cite{maze2018iarpa}. In LFW \cite{huang2008labeled}, CFP-FP \cite{sengupta2016frontal}, IJBB(Tar@Far=1e-4) \cite{whitelam2017iarpa}, IJBC (Tar@Far=1e-4) \cite{maze2018iarpa}, InterFace's best results is in the second rank. In IJBB(Tar@Far =1e-6) \cite{whitelam2017iarpa} and IJBC(Tar@Far=1e-6) \cite{maze2018iarpa}, InterFace's best results is in the third rank. In MegaFace(R)(Rank-1) \cite{kemelmacher2016megaface,deng2019arcface} and MegaFace(Rank-1) \cite{kemelmacher2016megaface}, InterFace's best results is in the fifth rank and in the fourth rank.

\begin{table*}
	\centering
	\scalebox{0.7}{
		\begin{tabular}{|c|c|c|c|c|c|c|c|}
			\hline
			\multirow{2}*{Method}&\multirow{2}*{Training Dataset} 			 & IJB-B(Tar@Far=1e-6)  &  IJB-B(Tar@Far=1e-5)  &IJB-B(Tar@Far=1e-4)    & IJB-C(Tar@Far=1e-6)   & IJB-C(Tar@Far=1e-5)  & IJB-C(Tar@Far=1e-4) \\
			\cline{3-8}
			~					 & ~				   					 	 & Accuracy(\%)  	& Accuracy(\%)	    &Accuracy(\%)       &Accuracy(\%) 		&Accuracy(\%)     	&Accuracy(\%) \\
			\hline
			Arcface \cite{deng2019arcface}
			&    MS1MV2 \cite{deng2019arcface,guo2016ms}
			& -		 		& -	    	   		& 94.2      		& -             	& -   				& 95.6 \\
			Dynamic-AdaCos \cite{jiao2021dyn}		 
			& clean MS1M \cite{guo2016ms,zhang2019adacos} + CASIA \cite{yi2014learning}
			& -	 			& -					& -					& 83.28		        & 88.03				& 92.4\\
			GroupFace \cite{kim2020groupface}
			& MS1MV2 \cite{deng2019arcface,guo2016ms}
			& \textbf{52.12}(1)& 91.24(2) 		 	& 94.93 			& \textbf{90.53} (1)& 94.53(3) 			& 96.26\\
			CircleLoss \cite{sun2020circle} 			
			& MS1MV2 \cite{deng2019arcface,guo2016ms}
			& -	 			& 					& - 				& - 				& 89.60 			& 93.95\\
			CurricularFace \cite{huang2020curricularface} 		 
			& MS1MV2 \cite{deng2019arcface,guo2016ms}
			& -				& -					& 94.8				& -					&-					& 96.1\\
			Partial-FC-ArcFace \cite{an2021partial} 	 
			& MS1MV2 \cite{deng2019arcface,guo2016ms}
			& -				& -	 				& 94.8 				& - 				& - 				& 96.2\\
			Partial-FC-CosFace \cite{an2021partial}   
			& MS1MV2 \cite{deng2019arcface,guo2016ms}
			& -	 			& -	 				& 95.0 				& - 				& -		 			& 96.4\\
			ElasticFace-Arc \cite{boutros2022elasticface}
			& MS1MV2 \cite{deng2019arcface,guo2016ms}
			& - 				& - 				& 95.22(3) 			& - 				& - 				& 96.49(3)\\
			ElasticFace-Cos \cite{boutros2022elasticface}	 
			&  MS1MV2 \cite{deng2019arcface,guo2016ms}
			& -				& -					& \textbf{95.43}(1) & - 				& -					& \textbf{96.65}(1)\\
			MagFace+ \cite{meng2021magface} 			 
			&    MS1MV2 \cite{deng2019arcface,guo2016ms}
			& 42.32	 		& 90.36				& 94.51 			& 90.24 (2)			& 94.08				& 95.97\\
			\hline
			Softmax* \cite{meng2021magface}	
			&    MS1MV2 \cite{deng2019arcface,guo2016ms} 
			& 46.73(2) 	 	& 75.17	    	    & 90.06     		& 64.07             & 83.68   			& 92.40\\
			SphereFace* \cite{liu2017sphereface,meng2021magface}
			& 	  MS1MV2 \cite{deng2019arcface,guo2016ms} 
			& 39.40	 		& 73.58				& 89.19				& 68.86		        & 83.33				& 91.77\\
			CosFace* \cite{wang2018cosface,meng2021magface}
			&    MS1MV2 \cite{deng2019arcface,guo2016ms} 
			& 40.41	        & 89.25			 	& 94.01				& 87.96		 		& 92.68				& 95.56\\
			ArcFace* \cite{deng2019arcface,meng2021magface}
			& 	  MS1MV2 \cite{deng2019arcface,guo2016ms} 
			& 38.68        	& 88.50				& 94.09			 	& 85.65		        & 92.69				& 95.74\\
			\hline
			InterFace$_{cid\&ct}$  & MS1MV2 \cite{deng2019arcface,guo2016ms} 
			& 45.43(3)	 		& 90.76(3)			& 95.17				& 90.08 (3)			& 94.81(2) 			& 96.54(2)\\
			InterFace$_{did\&ct}$ & MS1MV2 \cite{deng2019arcface,guo2016ms} 
			& 43.84	 		& 89.62				& 94.73				& 88.46 			& 94.34				& 96.29\\
			InterFace$_{did\&dt}$ & MS1MV2 \cite{deng2019arcface,guo2016ms}  			
			& 43.67	 		& \textbf{91.57}(1)	& 95.23(2)			& 89.68 			& \textbf{94.93}(1) & 96.49(3)\\
			\hline
		\end{tabular}
	}
	\caption{The achieved results on the IJB-B and IJB-C benchmarks. On IJB-C(Tar@Far=1e-5), InterFace solutions extend state-of-the-art performances. tiFace scores very close to the state-of-the-art on Other benchmarks."*" marks the reproduced code of other articles, the first quote is the original article, and the second quote is the reproduced article. The top performance in each benchmark is in bold. All decimal points are provided as reported in the respective works. The top three performances in each benchmark are noted with a rank number between parentheses(1,2 or 3).}
	\label{tab:ijb}
\end{table*}
\paragraph{Result on LFW, AgeDB-30, CALFW, CPLFW, and CFP-FP}
The models has been train under the parameters specified in Parameter Selection \ref{ps} and evaluated at dataset level LFW, AgeDB-30, CALFW, CPLFW, and CFP-FP. We present the results of the evaluation of our listed solutions in LFW \cite{huang2008labeled}, AgeDB-30 \cite{moschoglou2017agedb}, CALFW \cite{zheng2017cross}, CPLFW \cite{zheng2018cross} and CFP-FP \cite{sengupta2016frontal} in a Table.\ref{tab:lfw} divided into two parts, the top half being the evaluation results obtained for the previously worked solutions and the bottom half being the evaluation results obtained for our proposed solutions. As can be seen from Table.\ref{tab:lfw}, our evaluation results are more concerned with age-related datasets. These datasets are AgeDB-30 \cite{moschoglou2017agedb} and CALFW \cite{zheng2017cross} benchmarks. In the AgeDB-30 \cite{moschoglou2017agedb} dataset, the accuracy of our proposed InterFace$_{cid\&ct}$ solution and InterFace$_{did\&dt}$ solution in those dataset were 98.37\% and 98.38\% respectively,  while the accuracy of the top state-of-the-art performance was 98.35\% from the ElasticFace \cite{boutros2022elasticface} solution. In CALFW \cite{zheng2017cross}, the accuracy of our proposed InterFace$_{cid\&ct}$ solution and InterFace$_{did\&dt}$ solution in this dataset were 96.22\% and 96.27\% respectively, while the top state-of-the-art performance comes from the GroupFace \cite{kim2020groupface}, CircleLoss \cite{sun2020circle} and Part-FC-ArcFace \cite{an2021partial} solutions, with an accuracy of 96.20\%.
This significantly improves the performance in age-changing scenarios, and InterFace optimizes the distribution of the feature space to enhance the generalization ability of the model.
Compared with AgeDB-30 \cite{moschoglou2017agedb} and CALFW \cite{zheng2017cross}, CPLFW \cite{zheng2018cross} and CFP-FP \cite{sengupta2016frontal} pay more attention to the change of posture on the face.
On the CPLFW \cite{zheng2018cross} dataset, the InterFace$_{cid\&ct}$ solution was evaluated with an accuracy of 93.25\% which is very close to the top state-of-the-art accuracy from Elasticface-Arc \cite{boutros2022elasticface}, and the InterFace$_{did\&dt}$ solution was evaluated with an accuracy of 98.28\% and top state-of-the-art the same accuracy as art comes from the Elasticface-Arc \cite{boutros2022elasticface} solution. On the CFP-FP \cite{sengupta2016frontal} dataset, the InterFace$_{cid\&ct}$ solution and InterFace$_{did\&dt}$ solution are evaluated with 98.66\% and 98.5\% accuracies, respectively, very close to the top state-of-the-art accuracy of 99.73\% from the ElasticFace-Cos \cite{boutros2022elasticface} solution.
In the LFW \cite{huang2008labeled} benchmark, LFW \cite{huang2008labeled} is the earliest proposed Labeled Faces in the Wild dataset, many solutions are close to saturation in this dataset, our InterFace solution and InterFace solution are both evaluated with an accuracy of 99.83\%, which is very Close to the top state-of-the-art from the GroupFace \cite{kim2020groupface} solution with an accuracy of 99.85\%. For InterFace$_{dit\&ct}$, it may not show relatively good performance, but it enriches the integrity of the experiment.
\begin{table}
	\centering
	\scalebox{0.65}{
		\begin{tabular}{|c|c|c|c|c|c|c|c|}
			\hline
			\multirow{2}*{Method}& \multirow{2}*{Training Dataset} 			 & MegaFace(R)(Rank1)& MegaFace(Rank1)\\
			\cline{3-4}
			~					 &	~			   					 		 &Accuracy(\%)       &Accuracy(\%)     \\
			\hline
			Arcface \cite{deng2019arcface}
			&    MS1MV2 \cite{deng2019arcface,guo2016ms}
			& 98.35     	     & 81.03   		\\
			CosFace \cite{wang2018cosface}
			&    private      					 		 & -				 & \textbf{82.72}(1)	\\
			Dynamic-AdaCos \cite{zhang2019adacos}		 
			& clean MS1M \cite{guo2016ms,zhang2019adacos} + CASIA \cite{yi2014learning}
			& 97.41		 	 & -					\\
			AdaptiveFace \cite{liu2019adaptiveface} 		 
			&    clean MS1M \cite{deng2019arcface,wu2018light}
			& 95.02			 & -				\\
			UniformFace \cite{duan2019uniformface}
			& clean MS1M \cite{deng2019arcface,guo2016ms}  + VGGFace2 \cite{cao2018vggface2}
			& -			     & 79.98			\\
			GroupFace \cite{kim2020groupface}			 
			& MS1MV2 \cite{deng2019arcface,guo2016ms}
			& 98.74(2) 		 & 81.31(2) 		\\
			CircleLoss \cite{sun2020circle} 			 
			& MS1MV2 \cite{deng2019arcface,guo2016ms}
			& 98.50 			 & - 				\\
			CurricularFace \cite{huang2020curricularface}
			& MS1MV2 \cite{deng2019arcface,guo2016ms}
			& 98.71(3)			 & 81.26(3)		\\
			Dyn-arcFace \cite{jiao2021dyn} 		 
			& MS1MV2 \cite{deng2019arcface,guo2016ms}
			& - 				 & -	 	 		\\
			MagFace \cite{meng2021magface} 			 
			& MS1MV2 \cite{deng2019arcface,guo2016ms}
			& -		 	 	 & -		 			\\
			Partial-FC-ArcFace \cite{an2021partial} 	 
			& MS1MV2 \cite{deng2019arcface,guo2016ms}
			& 98.31 			 & - 				\\
			Partial-FC-CosFace \cite{an2021partial}   
			& MS1MV2 \cite{deng2019arcface,guo2016ms}
			& 98.36 			 & -		 			\\
			ElasticFace-Arc \cite{boutros2022elasticface}   	 
			& MS1MV2 \cite{deng2019arcface,guo2016ms}
			& \textbf{98.81(1)} & 80.76 		\\
			ElasticFace-Cos \cite{boutros2022elasticface}
			& MS1MV2 \cite{deng2019arcface,guo2016ms}
			& 98.70 			 & 81.08 			\\
			\hline
			InterFace$_{cid\&ct}$  & MS1MV2 \cite{deng2019arcface,guo2016ms}
			& 98.62			 & 81.01 		\\
			InterFace$_{did\&ct}$  & MS1MV2 \cite{deng2019arcface,guo2016ms}
			& 98.41			 & 81.03 		\\
			InterFace$_{did\&dt}$  & MS1MV2 \cite{deng2019arcface,guo2016ms}
			& 98.58			 & 81.21 		\\
			\hline
		\end{tabular}
	}
	\caption{ 
		The achieved results on the MegaFace benchmarks. MegaFace(R) is refined in [paper]. The top performance in each benchmark is in bold. All decimal points are provided as reported in the respective works.The top three performances in each benchmark are noted with rank number between parentheses(1,2 or 3).
	}
	\label{tab:mega}
\end{table}
\paragraph{Result on MegaFace}
In this part of the evaluation, we use the same trained models for the evaluations of MegaFace(R)(Rank-1) \cite{kemelmacher2016megaface} and MegaFace(Rank-1) \cite{kemelmacher2016megaface}. For the division of the structure of Table.\ref{tab:mega} we refer to the structure of Table.\ref{tab:lfw}. In the evaluation validation of MegaFace(R) (Rank-1) \cite{kemelmacher2016megaface} and MegaFace(Rank-1) \cite{kemelmacher2016megaface}, InterFace did not achieve the same effect as RESULTS.A section. In the Rank-1 evaluation criteria of the MegaFace(R) \cite{kemelmacher2016megaface} dataset, InterFace$_{cid\&ct}$, InterFace$_{did\&ct}$ and  InterFace$_{did\&dt}$ obtained 98.62\%, 98.41\% and 98.58\% accuracy, respectively. This score is very close to the performance of top-the-state-of-art. In the evaluation criteria of the MegaFace \cite{kemelmacher2016megaface} dataset Rank-1, InterFace$_{cid\&ct}$, InterFace$_{did\&ct}$ and  InterFace$_{did\&dt}$ obtained 81.01\%, 81.03\% and 81.21\% accuracy, respectively, which is also very close to the performance of top-the-state-of-art. 
\paragraph{Result on IJB-B,IJB-C}
In this section, we continue to maintain the consistency of the evaluation model. InterFace was evaluated for IJB-B(Tar@Far=1e-4,Tar@Far=1e-5,Tar@Far=1e-6) \cite{whitelam2017iarpa} and IJBC(Tar@Far=1e-4,Tar@Far=1e-5,Tar@Far=1e-6) \cite{maze2018iarpa}. In this section we have a different structure for Table.\ref{tab:ijb} than for Table.\ref{tab:lfw} and Table.\ref{tab:mega}, where Table.\ref{tab:ijb} is divided into 3 parts, the top part is the evaluation results of the previously worked solutions, the middle part is the evaluation results of other solutions replicated in MagFace \cite{meng2021magface} and marked with an "*", and the bottom part is the evaluation results of our current proposed solutions.
By observing Table.\ref{tab:ijb} we can find that InterFace$_{cid\&ct}$ and InterFace$_{did\&dt}$ have a better performance compared to InterFace$_{did\&ct}$ and other solutions.\par
InterFace$_{did\&dt}$ obtained 91.57\% and 94.93\% accuracy on IJBB (Tar@Far=1e-4) and IJBC (Tar@Far=1e-5), while top state-of-the-art performance in these two evaluations was 91.24\% accuracy from ElasticFace-Cos and 94.53\% accuracy from GroupFace. 
On IJBB (Tar@Far=1e-5) \cite{whitelam2017iarpa} and IJBC (Tar@Far=1e-5) \cite{maze2018iarpa}, InterFace$_{cid\&ct}$ obtained 90.76\%  accuracy that is rank-3 and 94.81\% accuracy that is rank-2.  
On IJBB (Tar@Far=1e-6) \cite{whitelam2017iarpa} and IJBC (Tar@Far=1e-6) \cite{maze2018iarpa}, InterFac$e_{cid\&ct}$ obtained an accuracy of 45.43\% and 90.08\% and InterFace$_{did\&dt}$ obtained an accuracy of 43.67\% and 89.68\%. In the above two validation criteria, InterFace$_{cid\&ct}$ obtained the rank-3 and InterFace$_{did\&dt}$ obtained the rank-4 in both in the ranking. 
On IJBB (Tar@Far=1e-4) \cite{whitelam2017iarpa} and IJBC (Tar@Far=1e-4) \cite{maze2018iarpa}, InterFace$_{cid\&ct}$ obtained 95.17\% and 96.45\% accuracy and InterFace$_{did\&dt}$ obtained 95.23\% and 96.49\% accuracy. In the above two validation criteria, InterFace$_{cid\&ct}$ obtained the rank-4 and second rank-2, respectively and InterFace$_{did\&dt}$ obtained the rank-2 and rank-3, respectively. 
Those evaluation results are very close to the performance of top state-of-the-art
For InterFace$_{dit\&ct}$, it may not show relatively good performance, but it enriches the integrity of the experiment.

\section{Conclusions}
In this paper, we propose InterFace Loss, which compensates for the inability to add different margin corrections to samples to other classes considering both inter-class and intra-class distributions. Our motivation is that the real sample distances to different class centers should be inconsistent, and also the intra-class variation of different samples in different classes is inconsistent, adding fixed margin penalties to different inter-classes on softmax is not consistent with the distribution of realistic data and cannot be a better model to learn better spatial distribution. Therefore, we introduce an intra-class ratio to make the samples add dynamic margin penalty to different classes in order to let the model learn better spatial distribution. We have evaluated our InterFace and top state-of-the-art face recognition solutions on 13 benchmarks for comparison. the InterFace solution continuously extends the state-of-the-art face recognition benchmarks (five out of thirteen). By comparing the final results we can conclude that InterFace has better generalization ability for age-related benchmarks.

\section*{Acknowledgment}
The authors thanks to the experimental platform provided by the Laboratory of Pattern Recognition and Artificial Intelligence, also thanks to the dataset for construction safety detection provided by Yunnan Power Grid Co Ltd Yuxi Power Supply Bureau. This work is supported by National Natural Science Foundation of China [41971392]; Yunnan Province Ten-thousand Talents Program.

{\small
\bibliographystyle{ieee}
\bibliography{main}

\begin{thebibliography}{10}\itemsep=-1pt

\bibitem{an2021partial}
X.~An, X.~Zhu, Y.~Gao, Y.~Xiao, Y.~Zhao, Z.~Feng, L.~Wu, B.~Qin, M.~Zhang,
  D.~Zhang, et~al.
\newblock Partial fc: Training 10 million identities on a single machine.
\newblock In {\em Proceedings of the IEEE/CVF International Conference on
  Computer Vision}, pages 1445--1449, 2021.

\bibitem{boutros2022elasticface}
F.~Boutros, N.~Damer, F.~Kirchbuchner, and A.~Kuijper.
\newblock Elasticface: Elastic margin loss for deep face recognition.
\newblock In {\em Proceedings of the IEEE/CVF Conference on Computer Vision and
  Pattern Recognition}, pages 1578--1587, 2022.

\bibitem{cao2018vggface2}
Q.~Cao, L.~Shen, W.~Xie, O.~M. Parkhi, and A.~Zisserman.
\newblock Vggface2: A dataset for recognising faces across pose and age.
\newblock In {\em 2018 13th IEEE international conference on automatic face \&
  gesture recognition (FG 2018)}, pages 67--74. IEEE, 2018.

\bibitem{chopra2005learning}
S.~Chopra, R.~Hadsell, and Y.~LeCun.
\newblock Learning a similarity metric discriminatively, with application to
  face verification.
\newblock In {\em 2005 IEEE Computer Society Conference on Computer Vision and
  Pattern Recognition (CVPR'05)}, volume~1, pages 539--546. IEEE, 2005.

\bibitem{deng2019arcface}
J.~Deng, J.~Guo, N.~Xue, and S.~Zafeiriou.
\newblock Arcface: Additive angular margin loss for deep face recognition.
\newblock In {\em Proceedings of the IEEE/CVF conference on computer vision and
  pattern recognition}, pages 4690--4699, 2019.

\bibitem{duan2019uniformface}
Y.~Duan, J.~Lu, and J.~Zhou.
\newblock Uniformface: Learning deep equidistributed representation for face
  recognition.
\newblock In {\em Proceedings of the IEEE/CVF Conference on Computer Vision and
  Pattern Recognition}, pages 3415--3424, 2019.

\bibitem{guo2016ms}
Y.~Guo, L.~Zhang, Y.~Hu, X.~He, and J.~Gao.
\newblock Ms-celeb-1m: A dataset and benchmark for large-scale face
  recognition.
\newblock In {\em European conference on computer vision}, pages 87--102.
  Springer, 2016.

\bibitem{he2016deep}
K.~He, X.~Zhang, S.~Ren, and J.~Sun.
\newblock Deep residual learning for image recognition.
\newblock In {\em Proceedings of the IEEE conference on computer vision and
  pattern recognition}, pages 770--778, 2016.

\bibitem{huang2008labeled}
G.~B. Huang, M.~Mattar, T.~Berg, and E.~Learned-Miller.
\newblock Labeled faces in the wild: A database forstudying face recognition in
  unconstrained environments.
\newblock In {\em Workshop on faces in'Real-Life'Images: detection, alignment,
  and recognition}, 2008.

\bibitem{huang2020curricularface}
Y.~Huang, Y.~Wang, Y.~Tai, X.~Liu, P.~Shen, S.~Li, J.~Li, and F.~Huang.
\newblock Curricularface: adaptive curriculum learning loss for deep face
  recognition.
\newblock In {\em proceedings of the IEEE/CVF conference on computer vision and
  pattern recognition}, pages 5901--5910, 2020.

\bibitem{jiao2021dyn}
J.~Jiao, W.~Liu, Y.~Mo, J.~Jiao, Z.~Deng, and X.~Chen.
\newblock Dyn-arcface: dynamic additive angular margin loss for deep face
  recognition.
\newblock {\em Multimedia Tools and Applications}, 80(17):25741--25756, 2021.

\bibitem{kemelmacher2016megaface}
I.~Kemelmacher-Shlizerman, S.~M. Seitz, D.~Miller, and E.~Brossard.
\newblock The megaface benchmark: 1 million faces for recognition at scale.
\newblock In {\em Proceedings of the IEEE conference on computer vision and
  pattern recognition}, pages 4873--4882, 2016.

\bibitem{kim2020groupface}
Y.~Kim, W.~Park, M.-C. Roh, and J.~Shin.
\newblock Groupface: Learning latent groups and constructing group-based
  representations for face recognition.
\newblock In {\em Proceedings of the IEEE/CVF Conference on Computer Vision and
  Pattern Recognition}, pages 5621--5630, 2020.

\bibitem{liu2019adaptiveface}
H.~Liu, X.~Zhu, Z.~Lei, and S.~Z. Li.
\newblock Adaptiveface: Adaptive margin and sampling for face recognition.
\newblock In {\em Proceedings of the IEEE/CVF Conference on Computer Vision and
  Pattern Recognition}, pages 11947--11956, 2019.

\bibitem{liu2017sphereface}
W.~Liu, Y.~Wen, Z.~Yu, M.~Li, B.~Raj, and L.~Song.
\newblock Sphereface: Deep hypersphere embedding for face recognition.
\newblock In {\em Proceedings of the IEEE conference on computer vision and
  pattern recognition}, pages 212--220, 2017.

\bibitem{liu2016large}
W.~Liu, Y.~Wen, Z.~Yu, and M.~Yang.
\newblock Large-margin softmax loss for convolutional neural networks.
\newblock {\em arXiv preprint arXiv:1612.02295}, 2016.

\bibitem{maze2018iarpa}
B.~Maze, J.~Adams, J.~A. Duncan, N.~Kalka, T.~Miller, C.~Otto, A.~K. Jain,
  W.~T. Niggel, J.~Anderson, J.~Cheney, et~al.
\newblock Iarpa janus benchmark-c: Face dataset and protocol.
\newblock In {\em 2018 international conference on biometrics (ICB)}, pages
  158--165. IEEE, 2018.

\bibitem{meng2021magface}
Q.~Meng, S.~Zhao, Z.~Huang, and F.~Zhou.
\newblock Magface: A universal representation for face recognition and quality
  assessment.
\newblock In {\em Proceedings of the IEEE/CVF Conference on Computer Vision and
  Pattern Recognition}, pages 14225--14234, 2021.

\bibitem{moschoglou2017agedb}
S.~Moschoglou, A.~Papaioannou, C.~Sagonas, J.~Deng, I.~Kotsia, and
  S.~Zafeiriou.
\newblock Agedb: the first manually collected, in-the-wild age database.
\newblock In {\em proceedings of the IEEE conference on computer vision and
  pattern recognition workshops}, pages 51--59, 2017.

\bibitem{schroff2015facenet}
F.~Schroff, D.~Kalenichenko, and J.~Philbin.
\newblock Facenet: A unified embedding for face recognition and clustering.
\newblock In {\em Proceedings of the IEEE conference on computer vision and
  pattern recognition}, pages 815--823, 2015.

\bibitem{sengupta2016frontal}
S.~Sengupta, J.-C. Chen, C.~Castillo, V.~M. Patel, R.~Chellappa, and D.~W.
  Jacobs.
\newblock Frontal to profile face verification in the wild.
\newblock In {\em 2016 IEEE winter conference on applications of computer
  vision (WACV)}, pages 1--9. IEEE, 2016.

\bibitem{sohn2016improved}
K.~Sohn.
\newblock Improved deep metric learning with multi-class n-pair loss objective.
\newblock {\em Advances in neural information processing systems}, 29, 2016.

\bibitem{sun2020circle}
Y.~Sun, C.~Cheng, Y.~Zhang, C.~Zhang, L.~Zheng, Z.~Wang, and Y.~Wei.
\newblock Circle loss: A unified perspective of pair similarity optimization.
\newblock In {\em Proceedings of the IEEE/CVF Conference on Computer Vision and
  Pattern Recognition}, pages 6398--6407, 2020.

\bibitem{wang2018cosface}
H.~Wang, Y.~Wang, Z.~Zhou, X.~Ji, D.~Gong, J.~Zhou, Z.~Li, and W.~Liu.
\newblock Cosface: Large margin cosine loss for deep face recognition.
\newblock In {\em Proceedings of the IEEE conference on computer vision and
  pattern recognition}, pages 5265--5274, 2018.

\bibitem{whitelam2017iarpa}
C.~Whitelam, E.~Taborsky, A.~Blanton, B.~Maze, J.~Adams, T.~Miller, N.~Kalka,
  A.~K. Jain, J.~A. Duncan, K.~Allen, et~al.
\newblock Iarpa janus benchmark-b face dataset.
\newblock In {\em proceedings of the IEEE conference on computer vision and
  pattern recognition workshops}, pages 90--98, 2017.

\bibitem{wu2018light}
X.~Wu, R.~He, Z.~Sun, and T.~Tan.
\newblock A light cnn for deep face representation with noisy labels.
\newblock {\em IEEE Transactions on Information Forensics and Security},
  13(11):2884--2896, 2018.

\bibitem{yi2014learning}
D.~Yi, Z.~Lei, S.~Liao, and S.~Z. Li.
\newblock Learning face representation from scratch.
\newblock {\em arXiv preprint arXiv:1411.7923}, 2014.

\bibitem{zhang2019adacos}
X.~Zhang, R.~Zhao, Y.~Qiao, X.~Wang, and H.~Li.
\newblock Adacos: Adaptively scaling cosine logits for effectively learning
  deep face representations.
\newblock In {\em Proceedings of the IEEE/CVF Conference on Computer Vision and
  Pattern Recognition}, pages 10823--10832, 2019.

\bibitem{zheng2018cross}
T.~Zheng and W.~Deng.
\newblock Cross-pose lfw: A database for studying cross-pose face recognition
  in unconstrained environments.
\newblock {\em Beijing University of Posts and Telecommunications, Tech. Rep},
  5:7, 2018.

\bibitem{zheng2017cross}
T.~Zheng, W.~Deng, and J.~Hu.
\newblock Cross-age lfw: A database for studying cross-age face recognition in
  unconstrained environments.
\newblock {\em arXiv preprint arXiv:1708.08197}, 2017.

\end{thebibliography}
}

\end{document}